\documentclass[letterpaper, 10 pt, conference]{ieeeconf}

\IEEEoverridecommandlockouts

\usepackage{booktabs}
\usepackage{array}
\usepackage[table]{xcolor}
\usepackage{url}
\usepackage{cite}
\usepackage{amsmath}
\usepackage{amssymb}
\usepackage{graphicx}
\makeatletter
\let\NAT@parse\undefined
\makeatother
\usepackage[hidelinks]{hyperref}
\hypersetup{
    pdftitle={EvoTER: Evolving Embodied Agents via Tool-Use Experience and Reflection},
    pdfauthor={Yingting Zhou, Yuyin Zhou, Yongliang Wang, Haoran Li, Dongbin Zhao}
}
\definecolor{evoRed}{HTML}{A65D56}
\definecolor{evoBlue}{HTML}{526F91}

\title{\LARGE \bf
EvoTER: Evolving Embodied Agents via Tool-Use \\ Experience and Reflection
}

\author{Yingting Zhou$^{1,2}$, Yuyin Zhou$^{1,2}$, Yongliang Wang$^{1,2,\dagger}$, Haoran Li$^{1,2}$, and Dongbin Zhao$^{1,2}$%
\thanks{$^{1}$The State Key Laboratory of Multimodal Artificial Intelligence Systems, Institute of Automation, Chinese Academy of Sciences, Beijing, China.}%
\thanks{$^{2}$University of Chinese Academy of Sciences, Beijing, China.}%
\thanks{$^{\dagger}$Corresponding author.}%
}

\begin{document}
\bstctlcite{ralBST}

\maketitle
\thispagestyle{empty}
\pagestyle{empty}

\begin{abstract}

Long-horizon embodied tasks require agents to ground instructions in changing observations, coordinate heterogeneous tools, and recover from failures that can propagate through downstream operations.
Although existing LLM-based agents retain experience and refine reusable skills, successful local repairs do not by themselves establish complete-task gains over the current shared procedure library.
We present EvoTER, an experience-driven framework that enables embodied agents to self-evolve tool-use skills without updating model parameters. 
EvoTER operates through a three-role evolution loop: Explorer records tool-use trajectories; 
Diagnoser distills successful and failed executions into skills for tool selection, 
parameter computation, postcondition checks, and recovery; 
and Validator retains candidate revisions only when they improve complete-task success in fresh sessions. 
In simulation, EvoTER improves overall task success from a pre-evolution baseline of $53.60\%$ to $80.40\%$. In real-world experiments, simulation initialization reduces the attempts required to acquire short-horizon skills, while continued real-world evolution improves long-horizon success from $20\%$ to $80\%$–$100\%$ across tasks and enables the evolving skill library to incorporate new tool functionality introduced after simulation development.
Project website: \mbox{\href{https://evoter-project.github.io/}{\textcolor{blue}{\textit{https://evoter-project.github.io/}}}}.

\end{abstract}


\section{INTRODUCTION}
\label{sec:intro}
Large language models (LLMs) enable embodied agents to translate natural-language instructions into executable programs over robotic APIs~\cite{liang2023codeaspolicies,fu2026capx}. By interleaving reasoning, tool calls, and execution feedback,  LLM-based agents coordinate perception, planning, and control for multi-step tasks~\cite{yao2023react}. Integrating pretrained visuomotor policies, including vision-language-action (VLA) models, further extends this repertoire by combining learned manipulation behaviors with conventional robotic tools~\cite{huang2026roboharness}. 
However, reliable long-horizon execution remains challenging, as errors in perception, tool coordination, and physical interaction can propagate across dependent steps.

These challenges motivate four complementary capabilities for reusable tool-use procedures: (i) \textbf{Grounding}: deriving tool arguments from current observations~\cite{liang2023codeaspolicies}; (ii) \textbf{Composition}: coordinating heterogeneous tools as subtasks and scene conditions change~\cite{huang2026roboharness}; (iii) \textbf{Verification}: checking physical outcomes and recovering from execution deviations~\cite{huang2023innermonologue}; and (iv) \textbf{Adaptability}: supporting reuse and refinement under changing physical conditions and tool interfaces~\cite{lu2026aspire}. These capabilities interact during execution, complicating procedure reuse across changing task states and environments.

Existing agents exploit execution feedback at different levels: closed-loop agents use it for execution and recovery~\cite{yao2023react,huang2023innermonologue}, but such corrections do not necessarily yield reusable knowledge. Memory-guided agents retain experience~\cite{harnessvla2026,huang2026roboharness}, while self-improving systems revise reusable skills~\cite{lu2026aspire,wen2026roborsi}. In tool-use settings, such revisions can modify tool selection, state-dependent parameter computation, sequencing, and recovery. Recent approaches assess revisions in different ways: transfer studies validate memory or stack updates before frozen target evaluation~\cite{he2026skill2real,chen2026recursia}, 
whereas feedback-driven systems use execution and regression checks during refinement~\cite{lu2026aspire,wen2026roborsi}. 
However, successful repairs or regression checks do not necessarily establish complete-task gains over the current shared library, as local changes may affect downstream operations. This issue becomes particularly relevant after sim-to-real transfer, where physical feedback can reveal new failures and motivate further revisions. With model parameters and tool implementations fixed within each evolution phase, 
we ask: \textit{How can agents select revisions based on complete-task gains over the current library to continually evolve reusable tool-use procedure libraries for long-horizon tasks across simulation and real-world refinement?}

To address this question, \textbf{EvoTER} evolves a persistent library of reusable, observation-grounded tool-use skills through feedback-driven revision and comparative complete-task acceptance. These skills support heterogeneous tool composition, postcondition checks, and recovery. The evolution loop comprises three roles sharing a frozen multimodal LLM: \textbf{Explorer} collects execution trajectories, \textbf{Diagnoser} identifies failures and proposes revisions, and \textbf{Validator} compares candidate and current libraries on matched complete-task runs. Only revisions improving aggregate complete-task validation success over the current library are retained. Model parameters, tool implementations, and evaluation settings remain fixed per evolution phase for controlled comparisons. Privileged simulation feedback may aid diagnosis, while skill execution and validation rollouts use only deployment-available observations and tools. Validated simulation skills initialize real-world evolution, where physical feedback triggers further revision-and-acceptance cycles. In simulation, EvoTER achieves a $26.80$ percentage-point gain in overall task success over its pre-evolution baseline. On real robots, simulation initialization reduces short-horizon skill-acquisition attempts, while continued evolution adapts long-horizon procedures to real-world conditions and tool interfaces. Our contributions are as follows:
\begin{itemize}
    \item We propose EvoTER, a framework organizing tool invocation, composition, outcome checks, and recovery into reusable, observation-grounded skills for long-horizon tasks without updating model parameters.

    \item We introduce an Explorer--Diagnoser--Validator loop that turns execution and simulation data into skills for real-world use, selects revisions by task-success gains, and continues evolving skills from real-world feedback.

    \item We demonstrate superior performance in simulation, and real-world experiments further confirm that simulation-initialized skills reduce development attempts while continued evolution supports long-horizon execution.
\end{itemize}

\section{RELATED WORK}

\label{sec:related_work}
\subsection{Tool Orchestration for Embodied Agents}
\label{sec:related_tools}
\begin{figure*}[t]
\centering
\includegraphics[width=\textwidth]{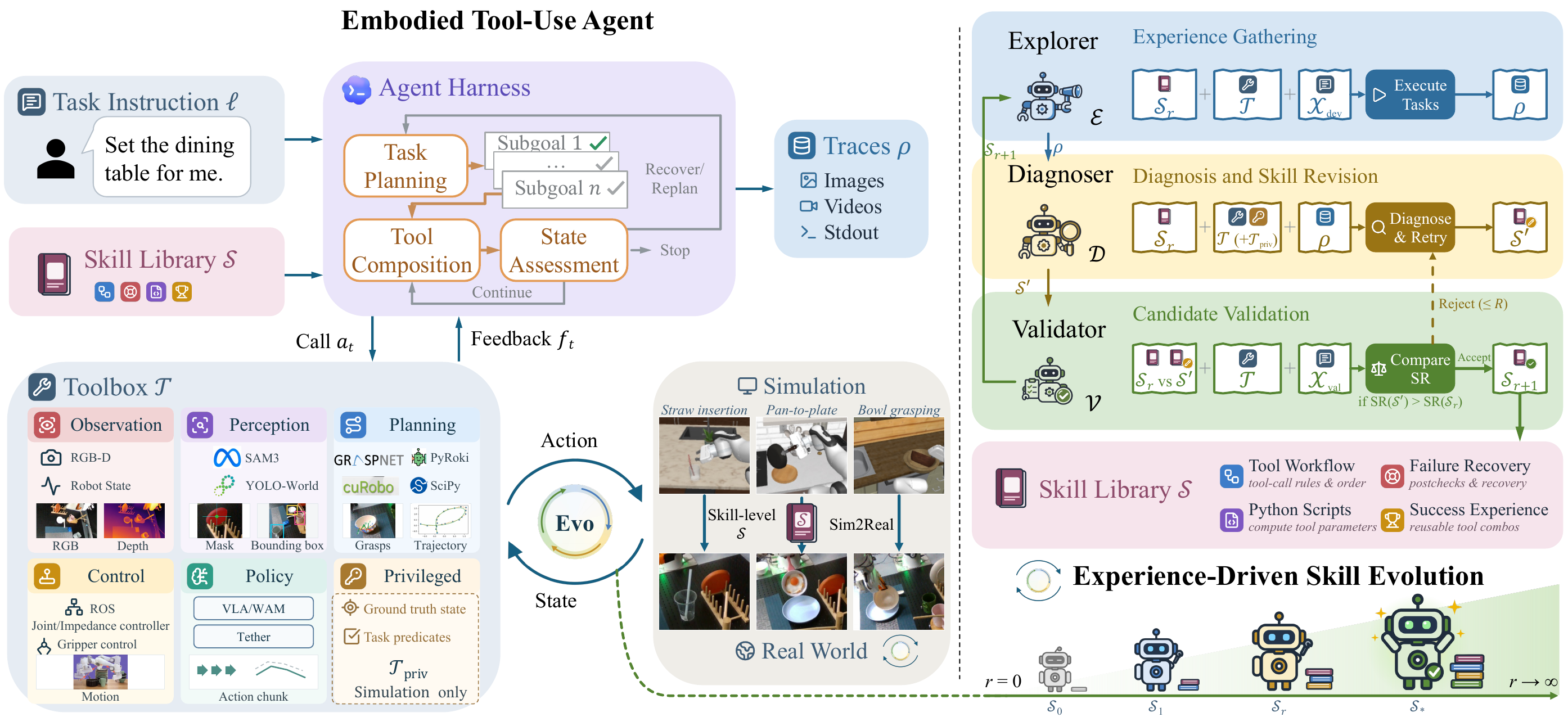}
\vspace{-6mm}
\caption{\textbf{Overview of EvoTER.} \textcolor{evoRed}{\textbf{\emph{Left: Embodied tool-use agent.}}} Skills guide the selection and composition of heterogeneous tools through a unified interface, while execution feedback closes the decision loop. \textcolor{evoBlue}{\textbf{\emph{Right: Experience-driven skill evolution.}}} \textcolor{cyan!60!black}{Explorer} collects execution trajectories; \textcolor{orange!60!black}{Diagnoser} extracts and revises skills, with privileged simulator information available only after exploration stagnates; \textcolor{green!60!black}{Validator} evaluates candidate and current libraries using ordinary observations and tools, retains candidates improving aggregate validation success, and returns rejected candidates to Diagnoser for revision.}

\label{fig:pipeline}
\vspace{-6mm}
\end{figure*}

Language-grounded planning connects natural-language instructions to executable robot skills through affordance-aware selection and environmental feedback~\cite{ly2026inteliplan,chen2025robogpt}. Programmatic approaches further make tool composition explicit through executable code over perception, planning, and control APIs~\cite{liang2023codeaspolicies}. Embodied coding frameworks extend this paradigm with reusable functions and refinable execution structures. CaP-X synthesizes task-agnostic functions from successful rollouts~\cite{fu2026capx}, while Graph-as-Policy refines computation graphs through simulation rehearsal~\cite{chen2026gap}. These approaches enable tool-use programs to be inspected, reused, and improved through execution.

Long-horizon tasks further require coordinating heterogeneous controllers with different capabilities and operating conditions. Successful tool calls may still leave physical states unsuitable for subsequent steps. Memory-guided harnesses address this challenge through execution history and adaptive orchestration. Harness VLA combines a frozen VLA with analytic primitives, using task traces and global memory to guide invocation and recovery~\cite{harnessvla2026}. RoboHarness characterizes policy capability boundaries and prepares handoff states, while supporting revisions to execution interfaces and orchestration~\cite{huang2026roboharness}. Despite these advances, improvements in tool selection, execution graphs, and policy handoffs do not alone establish whether reusable procedures benefit complete-task execution. EvoTER addresses this question by evolving observation-grounded procedures for tool invocation, composition, outcome checks, and recovery, with revisions selected through matched complete-task validation.

\subsection{Self-Improving Agents and Skill Transfer}
\label{sec}
Experience-driven agents retain successful programs and distill execution feedback into reusable guidance~\cite{meng2026lyra,li2025selu}. Dynamic skill banks organize reusable task- and step-level knowledge~\cite{tu2026d2skill}, while embodied skill-discovery methods acquire executable routines and diagnostic lessons through interaction~\cite{li2026embodiedevo}. Accumulating experience does not identify which procedures should change. Self-improving systems address this through fine-grained debugging and validated repairs~\cite{lu2026aspire}, hierarchical failure attribution~\cite{wen2026roborsi}, or evolving runtime critics, recovery strategies, and tools around frozen policies~\cite{zetta2026}. These approaches enable persistent improvement but differ in their revision scope and validation criteria.

Validation and transfer further determine whether acquired knowledge remains useful beyond development. Skill2Real transfers validated simulation skills as frozen memories~\cite{he2026skill2real}, while Recursia validates a broader executable stack before frozen target-task evaluation~\cite{chen2026recursia}. Complementary frameworks provide matched evaluation controls~\cite{wang2026sear}, session-level verification, and persistent execution records~\cite{phyagentos2026}. Real-world adaptation has also been explored: ASPIRE uses simulation skills to guide physical debugging and supports online skill learning~\cite{lu2026aspire}. Despite these advances, validating isolated repairs or frozen execution stacks does not alone establish how a shared skill library improves complete-task outcomes and continues adapting after transfer. Drawing on the distinction between privileged learning signals and deployment inputs, EvoTER separates privileged diagnosis from ordinary skill execution and validation. With model and tool implementations fixed within each evolution phase, EvoTER retains only skill-library revisions that improve matched complete-task success and supports continued real-world evolution as new tools become available.

\section{METHOD}
\label{sec:method}

EvoTER updates an embodied agent's skill library by converting tool-use experience into reusable skills through reflection and validation.
The framework comprises an embodied tool-use agent (\mbox{Sec.~\ref{sec:execution_system}}), an experience-driven skill evolution loop (\mbox{Sec.~\ref{sec:self_evolution}}), and a sim-to-real skill transfer and adaptation process (\mbox{Sec.~\ref{sec:sim_to_real}}).
Fig.~\ref{fig:pipeline} presents the framework overview, with each component detailed below.

\subsection{Problem Formulation}
\label{sec:problem_setup}

\noindent\textbf{Embodied Agent and Skill Library.}
We consider embodied agents that execute natural-language task
instructions through tool calls.
On platform $p\in\{\mathrm{sim},\mathrm{real}\}$, we define the agent
at evolution iteration $r$ as
\begin{equation}
\label{eq:agent_state}
    \mathcal{A}^{(p)}_r
    =
    \left(
        M,\mathcal{T}^{(p)},h^{(p)},\mathcal{S}^{(p)}_r
    \right),
\end{equation}
where $M$ is the model backbone, $\mathcal{T}^{(p)}$ the ordinary tool registry, $h^{(p)}$ the execution harness configuration, and $\mathcal{S}^{(p)}_r$ the persistent skill library. Each skill specifies applicability conditions, tool-call rules, parameter computation, postcondition checks, and recovery instructions, with optional executable scripts. Within each evolution phase, model parameters, ordinary tool implementations, and harness configuration remain fixed; the agent evolves by updating $\mathcal{S}^{(p)}_r$.

\noindent\textbf{Skill Evolver.}
A skill evolver converts the agent's execution experience into
persistent updates to its skill library.
Let $\mathcal{B}_r$ denote execution evidence collected by
$\mathcal{A}^{(p)}_r$ on development instances.
The evolver $\mathcal{U}$ proposes and validates skill updates
to produce the next library:
\begin{equation}
\label{eq:skill_evolution}
    \mathcal{S}^{(p)}_{r+1}
    =
    \mathcal{U}\!\left(
        \mathcal{S}^{(p)}_r,
        \mathcal{B}_r;
        \mathcal{X}_{\mathrm{val}}
    \right),
\end{equation}
where $\mathcal{X}_{\mathrm{val}}$ is a reserved validation set for assessing candidate updates. The operator includes update proposal and acceptance; rejected updates leave the library unchanged. Accepted updates persist for subsequent execution and evolution. EvoTER implements this process through three roles, Explorer, Diagnoser, and Validator, detailed in \mbox{Sec.~\ref{sec:self_evolution}}.
\noindent\textbf{Evolution Objective and Evaluation.}
Task instances are indexed by $(\kappa,\xi)$, where $\kappa$
denotes the task and $\xi$ its initial scene configuration.
Let $\mathcal{X}_{\mathrm{dev}}$ and $\mathcal{X}_{\mathrm{test}}$ be disjoint sets of task--seed pairs, with $\mathcal{X}_{\mathrm{val}}\subset\mathcal{X}_{\mathrm{dev}}$ reserved for intermediate validation.
For a fixed platform, let $\pi_M$ be $M$'s policy and
$y(\kappa,\xi;\pi_M,\mathcal{S})\in\{0,1\}$
the externally evaluated task outcome.
The task success rate (SR) on $\mathcal{X}$ is
\begin{equation}
\label{eq:success_rate}
    \mathrm{SR}(\mathcal{S};\mathcal{X})
    =
    \frac{1}{|\mathcal{X}|}
    \sum_{(\kappa,\xi)\in\mathcal{X}}
    y(\kappa,\xi;\pi_M,\mathcal{S}).
\end{equation}
We aim to convert execution experience into reusable skills that improve task success under the fixed execution system. Candidate acceptance depends on success rate on $\mathcal{X}_{\mathrm{val}}$. After development, the skill library is frozen and evaluated on $\mathcal{X}_{\mathrm{test}}$, which informs neither skill updating nor selection.

\subsection{Embodied Tool-Use Agent}
\label{sec:execution_system}

We formulate embodied execution as the composition of fixed tool capabilities guided by an evolving skill library. Tools provide the agent's executable capabilities, while skills encode procedural knowledge about their use across task states. The frozen model combines this knowledge with the task instruction and interaction history to select tools, determine their arguments, and assess execution outcomes. This separation allows skill updates to change the agent's execution behavior without modifying the model, tool implementations, or harness configuration.

\noindent\textbf{Fixed Tool Capabilities.}
The registry $\mathcal{T}^{(p)}$ exposes observation, perception, planning, control, and policy capabilities through tool interfaces, including open-vocabulary segmentation and detection~\cite{carion2026sam3,cheng2024yoloworld}, grasp generation~\cite{fang2023anygrasp}, and collision-free motion planning~\cite{sundaralingam2023curobo}. Each tool specifies its purpose and input--output schemas for model invocation. Simulation and real-robot platforms expose corresponding tool types with platform-specific implementations. The privileged diagnostic interface is separate from $\mathcal{T}^{(p)}$ (\mbox{Sec.~\ref{sec:self_evolution}}).

\begin{table*}[!t]
\vspace*{6pt}
\centering
\makeatletter
\long\def\@makecaption#1#2{%
  {\centering\normalfont\footnotesize
   #1.\enspace #2\par}%
  \vskip 3pt
}
\makeatother
\caption{\normalfont\textbf{Success rates on RoboCasa365 Target50.} }
\label{tab:main}
\begingroup
\scriptsize
\setlength{\tabcolsep}{4pt}
\renewcommand{\arraystretch}{1.0}
\begin{tabular*}{\textwidth}{@{\extracolsep{\fill}}l*{4}{c}@{}}
\toprule
& \multicolumn{4}{c}{Success rate (\%) $\uparrow$} \\
\cmidrule(lr){2-5}
Method & Atomic & Composite-Seen & Composite-Unseen & Overall \\
\midrule
\rowcolor{black!8}[0pt][0pt]
\multicolumn{5}{@{}l@{}}{\textit{Policy-based methods}} \\
WorldDreamer~\cite{robocasa365_leaderboard} & $66.30$ & $26.70$ & $9.00$ & $35.30$ \\
RLDX-1~\cite{robocasa365_leaderboard} & $67.60$ & $27.90$ & $8.50$ & $36.00$ \\
CheckVLA~\cite{checkvla2026} & $63.70$ & $30.90$ & $10.20$ & $36.10$ \\
Xiaomi-Robotics-1~\cite{xiaomi2026} & $80.20$ & $57.10$ & $32.10$ & $57.40$ \\
\midrule
\rowcolor{black!8}[0pt][0pt]
\multicolumn{5}{@{}l@{}}{\textit{Agent-based methods}} \\
Zetta~$\zeta$~\cite{zetta2026} & $93.56$ & -- & -- & -- \\
Harness VLA (Codex / GPT-6 Astra)~\cite{harnessvla2026} & $87.78$ & $43.75$ & $42.50$ & $59.20$ \\
HexaAnything (HexaModel v0.1 + XR-1)\textsuperscript{a}~\cite{hexaanything2026} & $81.00$ & $62.30$ & $39.50$ & $61.70$ \\
ProactiveVLA~\cite{proactivevla2026} & -- & $41.30$ & -- & -- \\
PhyAgentOS (RLDX-1)~\cite{phyagentos2026} & $75.60$ & \multicolumn{2}{c}{$24.40$\rlap{$^{\dagger}$}} & $42.80$ \\
EvoTER w/o skills ($\mathcal{S}_0$) & $81.11$ & $47.50$ & $28.75$ & $53.60$ \\
\textbf{EvoTER (ours)} & $\mathbf{94.44}$ & $\mathbf{78.75}$ & $\mathbf{66.25}$ & $\mathbf{80.40}$ \\
\bottomrule
\end{tabular*}
\par\smallskip
\begin{minipage}{\textwidth}
\scriptsize

Our Overall is the task-macro average over all $50$ tasks
($18$ Atomic, $16$ Composite-Seen, and $16$ Composite-Unseen).
Results for other methods are taken from the cited sources, retaining their reported rounding;
``--'' denotes an unreported subset or an unavailable full-benchmark result.
\textsuperscript{a}\,Uses the specially trained HexaModel v0.1.
$^{\dagger}$\,Reports the combined success rate over Composite-Seen and Composite-Unseen.
\end{minipage}
\endgroup
\vspace{-16pt}
\end{table*}

\noindent\textbf{Reusable Procedural Knowledge.}
The skill library $\mathcal{S}^{(p)}_r$ stores procedural knowledge
that guides tool selection and composition.
Each skill specifies when it applies, how to invoke tools and compute
their arguments, check postconditions, and recover
from unsuccessful execution.
Skill summaries support selection, and loaded skills provide
detailed execution guidance.
Associated scripts may compute observation-grounded tool arguments from current
non-privileged observations using ordinary tool interfaces.
The library remains fixed within an episode.
When $\mathcal{S}^{(p)}_0=\varnothing$, the model acts directly
from the tool descriptions.

\noindent\textbf{Feedback-Driven Tool Composition.}
At execution step $t$ of evolution iteration $r$, the frozen model
selects a tool call conditioned on the instruction $\ell_{\kappa}$,
observation $o_t$, history $\mathcal{H}_t$, and current skill library:
\begin{equation}
\label{eq:tool_policy}
    a_t=(\tau_{k_t},\phi_t)
    \sim
    \pi_M\!\left(
        \cdot \mid
        \ell_{\kappa},o_t,\mathcal{H}_t,\mathcal{S}^{(p)}_r
    \right),
\end{equation}
where $\tau_{k_t}\in\mathcal{T}^{(p)}$ and $\phi_t$ denotes its arguments. The harness executes the call and adds feedback $f_t$ and observation $o_{t+1}$ to the interaction history. The agent uses this evidence to check the active skill's postconditions, proceeding if satisfied; otherwise, it follows recovery guidance or replans from the current state. Observed outcomes thus guide tool composition throughout execution. Episodes end upon task completion, terminal failure, or tool-call budget exhaustion.

Each episode produces an execution record that links tool decisions
to their observed consequences:
\begin{equation}
\label{eq:execution_trace}
    \rho =
    \left(
        \{(o_t,a_t,f_t,o_{t+1})\}_{t=0}^{L-1},
        y
    \right),
    \qquad L\leq H,
\end{equation}
where $L$ is the number of tool calls, $H$ is the call budget, and $y$ is the externally evaluated task outcome. These records and associated images, videos, and logs support skill diagnosis, refinement, and validation.

\subsection{Experience-Driven Skill Evolution}
\label{sec:self_evolution}

Skill evolution requires extracting reusable execution rules from interaction trajectories and evaluating their effectiveness under deployment conditions. EvoTER organizes this process into three roles: Explorer $\mathcal{E}$ collects execution experience, Diagnoser $\mathcal{D}$ analyzes execution and revises skills,
and Validator $\mathcal{V}$ evaluates candidate updates. The roles share the same frozen model but operate with distinct input contexts and tool permissions. Each role transition initiates a new session with the corresponding context and permissions.
Execution trajectories, diagnostic records, and skill versions are exchanged through external files, while the skill library retains deployment knowledge accumulated across tasks.

\noindent\textbf{\textcolor{cyan!60!black}{Explorer.}}
Successful trajectories reveal reusable tool combinations and parameter choices, whereas failed trajectories help identify limitations in the current execution rules. Explorer collects both types of experience during task execution. For a development instance $(\kappa,\xi)\in\mathcal{X}_{\mathrm{dev}}\setminus\mathcal{X}_{\mathrm{val}}$, Explorer starts a new session and executes the task using the current skill library and ordinary tool registry $\mathcal{T}^{(p)}$. The execution procedure matches that used for evaluation with the skill library frozen. After each task, the recorded tool calls, execution feedback, images, and videos are passed to Diagnoser for skill extraction and failure diagnosis.

\noindent\textbf{\textcolor{orange!60!black}{Diagnoser.}}
Task outcomes alone do not reveal which tool calls or execution rules require revision. Diagnoser examines execution trajectories, the current skill library, and diagnostic records to relate tool calls to observed environmental changes. From successful executions, it extracts reusable tool combinations, parameter computation methods, and postcondition checks. For failed executions, it uses logs, tool outputs, keyframes, and videos to locate execution failures. Diagnostic retries test alternative invocation parameters and recovery rules, providing evidence for candidate revisions.

When a task makes no progress over consecutive evolution iterations and reaches a predefined stagnation threshold, simulation diagnosis accesses $\mathcal{T}_{\mathrm{priv}}$. This interface provides simulation state, task predicates, and privileged planning tools for failure analysis and recovery testing. Additional information from diagnosis must be converted into rules executable under deployment conditions.

The skill-generation prompts require revisions to be grounded in currently observable tool outputs and associated scripts to compute invocation parameters from current observations. These prompts also prohibit embedding privileged coordinates, simulator
object identifiers, or seed-specific constants directly in skills.
Diagnoser produces a diagnostic report $d$ and a skill update $\Delta\mathcal{S}$. Based on execution evidence and the current library, it decides whether to refine an existing skill or add a new entry, forming a candidate library:
\begin{equation}
\label{eq:diagnoser_update}
\begin{aligned}
(d,\Delta\mathcal{S})
&=\mathcal{D}\!\left(\{\rho_i\}_{i=1}^{N},\mathcal{S},\mathcal{T}_{\mathrm{diag}}^{(p)}\right),\\
\mathcal{S}'&=\mathcal{S}\oplus\Delta\mathcal{S}.
\end{aligned}
\end{equation}
Here, $\mathcal{S}=\mathcal{S}^{(p)}_r$ is the current library, and $\mathcal{S}'$ is the candidate awaiting validation. The set $\{\rho_i\}_{i=1}^{N}\subseteq\mathcal{B}_r$ contains the $N$ input trajectories, and $\oplus$ denotes the skill-update application. The diagnostic tool set $\mathcal{T}_{\mathrm{diag}}^{(p)}$ normally equals the ordinary tool registry. It includes the privileged interface only during simulation diagnosis when the stagnation condition is met.

\noindent\textbf{\textcolor{green!60!black}{Validator.}}
Diagnostic retries may use information unavailable during deployment. Validator therefore reevaluates candidate skills using only ordinary execution interfaces.

It compares current and candidate libraries on $\mathcal{X}_{\mathrm{val}}$ in fresh sessions with matched instances and ordinary tools. Acceptance requires higher success rates:
\begin{equation}
\label{eq:validator_accept}
\operatorname{accept}(\mathcal{S}')
=\mathbf{1}\!\left[
\mathrm{SR}(\mathcal{S}';\mathcal{X}_{\mathrm{val}})
>
\mathrm{SR}(\mathcal{S};\mathcal{X}_{\mathrm{val}})
\right].
\end{equation}
An accepted candidate becomes $\mathcal{S}^{(p)}_{r+1}$ for subsequent execution and exploration. Rejected candidates leave the current library unchanged and return with their validation trajectories to Diagnoser for refinement, with at most $R$ revisions.

\subsection{Sim-to-Real Skill Transfer and Adaptation}
\label{sec:sim_to_real}

EvoTER transfers simulation experience through its persistent skill library, providing initial procedural knowledge for real-world evolution. The frozen simulation library $\mathcal{S}_{*}^{(\mathrm{sim})}$ initializes the real-robot agent with previously acquired tool-use rules, postcondition checks, and recovery guidance. The subscript $*$ denotes a library frozen after its evolution phase. These skills are subsequently revised using real-world execution experience through the same exploration, diagnosis, and validation loop. Diagnoser adapts platform-dependent skill bindings and refines execution rules to accommodate real observations and manipulation conditions. Validator assesses candidate revisions using only ordinary real-robot tools, retaining accepted updates for subsequent iterations. Throughout, model parameters, real-robot tool implementations, and harness configuration remain fixed. The resulting real-world library is frozen as $\mathcal{S}_{*}^{(\mathrm{real})}$.

\section{EXPERIMENTS}
\label{sec:experiments}
We empirically evaluate whether experience-driven skill evolution improves task completion in EvoTER, with model parameters and ordinary-tool implementations fixed within each evolution phase. After detailing the experimental setup (\mbox{Sec.~\ref{sec:exp_setup}}), we evaluate the evolved skill library on atomic and composite tasks in simulation (\mbox{Sec.~\ref{sec:main_results}}) and analyze its dependence on tool availability during execution (\mbox{Sec.~\ref{sec:ablation}}). On the real robot, we examine the effect of simulation initialization on short-horizon skill evolution and the contribution of real-world feedback to long-horizon execution, including skill-library evolution to accommodate real-world tool interfaces and capabilities (\mbox{Sec.~\ref{sec:real}}).

\subsection{Experimental Setup}
\label{sec:exp_setup}

\noindent\textbf{Platforms and Tasks.}
In simulation, we use RoboCasa365 Target50~\cite{robocasa365} with the PandaOmron mobile manipulator. The benchmark comprises $18$ Atomic, $16$ Composite-Seen, and $16$ Composite-Unseen tasks. The Seen/Unseen labels indicate task exposure during VLA training; development episodes from Composite-Unseen tasks may contribute to skill evolution. On the real robot, we use a Franka arm with a two-finger soft gripper to evaluate short- and long-horizon skill evolution. The task definitions and comparison protocols are detailed in \mbox{Sec.~\ref{sec:real}}.

\noindent\textbf{Agent and Tool Configuration.}
To evaluate skill evolution under fixed execution capabilities, we keep model parameters and ordinary-tool implementations unchanged within each evolution phase. Both platforms use a Codex-based agent framework with \texttt{gpt-6.1-sol} at medium reasoning effort. Observation, perception, planning, control, and VLA tools are implemented separately for each platform. The simulation VLA tool uses Xiaomi-Robotics-1~\cite{xiaomi2026}, while its real-robot counterpart uses our trained $\pi_{0.5}$-based policy~\cite{black2025pi05}. Real-world execution uses SAM3~\cite{carion2026sam3} segmentation, YOLO-World~\cite{cheng2024yoloworld} detection, AnyGrasp~\cite{fang2023anygrasp} grasp generation, CuRobo~\cite{sundaralingam2023curobo} motion planning, low-level control, and the frozen VLA and Tether~\cite{liang2026tether} policies. Comparisons hold agent configuration, policy checkpoints, tool implementations, camera access, and execution constraints fixed.

\begin{figure}[!b]
\vspace{-4pt}
\centering
\includegraphics[width=\columnwidth]{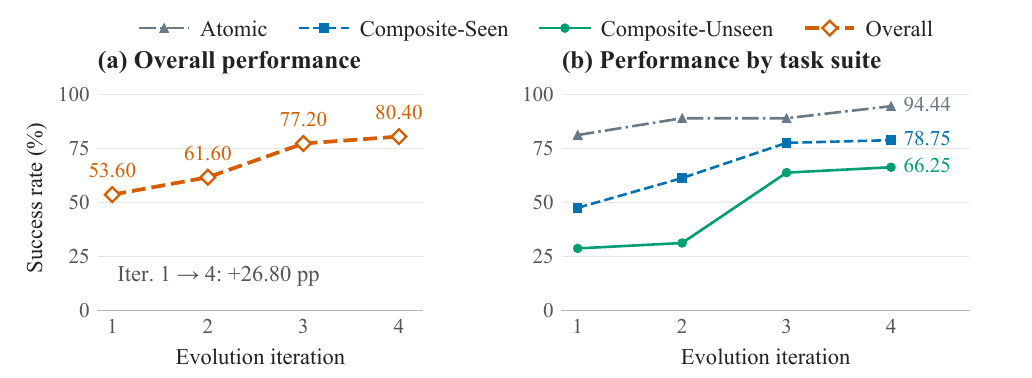}
\vspace{-20pt}
\caption{\textbf{Test success rates across evolution iterations.}
Overall and frozen-test SR on each RoboCasa365 Target50 subset at each evolution iteration; iteration~$1$ corresponds to evaluation with the initial empty library $\mathcal{S}_0$.}
\label{fig:evo_iter}
\vspace{-2mm}
\end{figure}

\noindent\textbf{Simulation Evolution Protocol.}
Starting from an empty skill library, we perform $3$ evolution rounds.
Each exploration round covers $50$ tasks with $5$ seeds per task, allowing an initial attempt and at most $2$ retries per task--seed instance.
After each round, Validator evaluates the current and candidate libraries on the same $5$ validation seeds per task, distinct from the exploration seeds.
Validation seeds change across rounds, with $\mathcal{X}_{\mathrm{val}}$ denoting the current round's validation instances.
Each library remains fixed during evaluation; the candidate is retained only if its SR exceeds that of the current library.
Validation records inform subsequent diagnosis and skill revision.
Fig.~\ref{fig:evo_iter} shows the initial empty-library evaluation followed by evaluations after the $3$ evolution rounds.

\noindent\textbf{Baselines.}
Our controlled comparison evaluates EvoTER against an empty-library agent under the same reasoning model, ordinary tools, and execution configuration, isolating the evolved skill library's contribution. For context, Table~\ref{tab:main} includes externally reported policy-based and agent-based methods, retaining their original evaluation protocols.

\noindent\textbf{Evaluation Metrics.}
We measure task completion using task-level success rate,
following Eq.~\eqref{eq:success_rate}. To weight tasks equally, we compute subset and Overall scores as unweighted means of the corresponding task-level success rates. For real-robot experiments, we additionally report skill acquisition costs. The cost definitions and task-specific success criteria are provided in \mbox{Sec.~\ref{sec:real}}.

\begin{figure*}[!t]
\centering
\includegraphics[width=\textwidth]{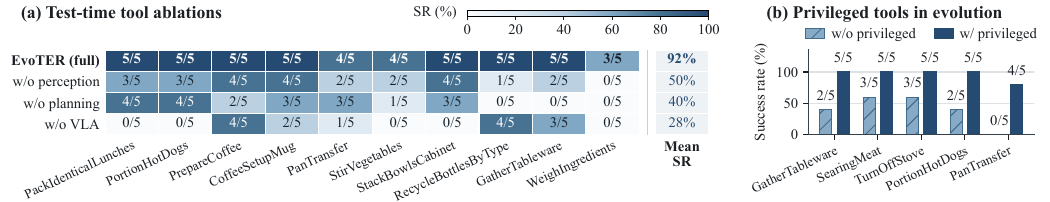}
\vspace{-6mm}
\caption{\textbf{{Test-time tool dependence and privileged diagnosis.}}
\textbf{(a)} Tool ablations on $10$ tasks. Cells show success counts ($k/5$), and the rightmost Mean SR column gives the unweighted mean of task-level success rates.
\textbf{(b)} Privileged diagnosis on five tasks with validation SR below $20\%$ until the stagnation threshold. From the same library at this threshold, evolution uses equal iteration budgets with or without added privileged simulation tools during diagnosis. Bars report frozen-test success over five trials per task; skill execution and test evaluation use only ordinary tools in both conditions.}

\label{fig:ablations_joint}
\label{fig:tool_ablation_tasks}
\label{fig:privileged}
\vspace{-2mm}
\end{figure*}

\begin{figure*}[!t]
\centering
\includegraphics[width=\textwidth]{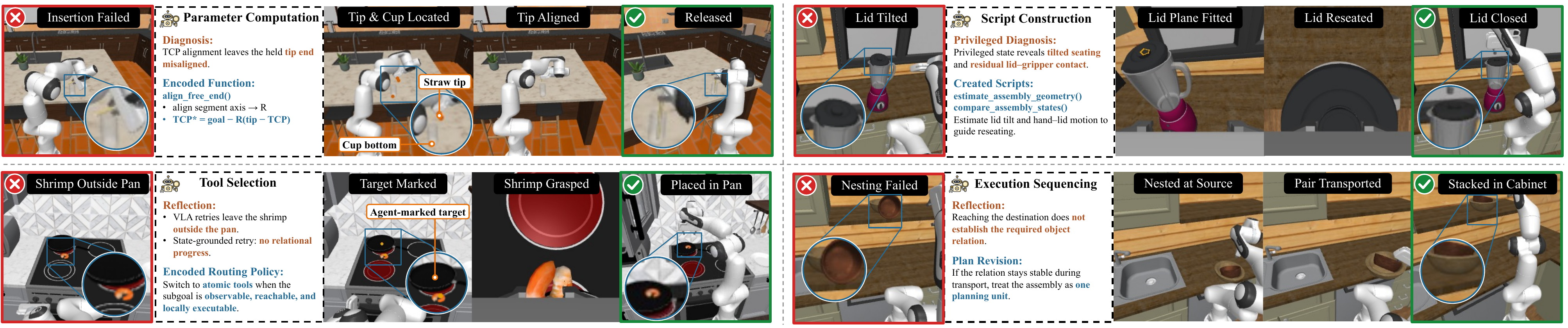}
\vspace{-6mm}
\caption{\textbf{Qualitative examples of skill revision in simulation.}
Each example depicts an initial failure (red), the corresponding diagnosis and skill update (dashed box), and the recovered execution (green).
\textbf{Parameter computation} (DeliverStraw): the tool center point (TCP) target is computed from the observed straw axis to place the free tip at the goal.
\textbf{Script construction} (CloseBlenderLid): privileged diagnosis of a tilted lid is distilled into two deployment-compatible RGB-D scripts.
\textbf{Tool selection} (SearingMeat): a routing rule replaces repeated VLA calls with atomic tools.
\textbf{Execution sequencing} (StackBowlsCabinet): the bowls are nested at the source and transported as a single planning unit.}
\label{fig:Four_Cases}
\vspace{-6mm}
\end{figure*}

\subsection{Skill Evolution Analysis}
\label{sec:main_results}
\noindent\textbf{Skill evolution improves execution under fixed capabilities.}
With the reasoning model, ordinary tools, and execution configuration held fixed, EvoTER increases Overall SR from $53.60\%$ to $80.40\%$ over the empty-library agent, yielding a gain of $26.80$ percentage points (pp) (Table~\ref{tab:main}).
This controlled comparison attributes the improvement to the evolved skill library under matched execution conditions.
The library remains frozen throughout testing, showing that the accumulated execution knowledge benefits subsequent task instances without further skill updates.
The result thus supports the central premise of EvoTER: execution experience can improve an embodied agent through persistent procedural knowledge while its underlying model and tool capabilities remain unchanged.
Across evolution iterations, Overall SR rises from $53.60\%$ at the first iteration, which corresponds to the empty library, to $80.40\%$ at the fourth (Fig.~\ref{fig:evo_iter}), with the largest single-iteration gain on Composite-Unseen between the second and third iterations.

\noindent\textbf{Composite tasks exhibit larger absolute gains.}
The improvement over the empty-library agent reaches $31.25$~pp on Composite-Seen and $37.50$~pp on Composite-Unseen, compared with $13.33$~pp on Atomic.
Because success requires completing the entire task, these gains reflect more successful end-to-end execution across the constituent stages.
This pattern is consistent with reusable tool-use procedures supporting the coordination of tool selection, parameter computation, and intermediate outcome checks in composite tasks.
Nevertheless, Composite-Unseen SR remains lower than Composite-Seen SR ($66.25\%$ versus $78.75\%$), indicating that substantial performance differences between the two subsets persist after skill evolution.

\noindent\textbf{Privileged simulation diagnosis helps resolve stagnated tasks.}
Fig.~\ref{fig:privileged}(b) compares skill evolution with and without the privileged diagnostic interface on five simulation tasks. With privileged diagnosis, the libraries achieve $5/5$ on four tasks and $4/5$ on PanTransfer, versus $0/5$ to $3/5$ without it (mean SR $96\%$ versus $40\%$). Because skills and validation use only ordinary tools, this gain reflects diagnostic information converted into deployment-executable procedures.

\noindent\textbf{Skill revisions take several forms.}
Fig.~\ref{fig:Four_Cases} links observed failures to revised tool-use rules.
Geometry-based revisions correct insertion targets by grounding them in current observations,
while execution feedback motivates changes in tool choice or execution order. The resulting procedures guide subsequent execution through the updated skill library.

\subsection{Tool Dependence Analysis}
\label{sec:ablation}

\begin{table*}[!t]
\vspace*{6pt}
\centering
\makeatletter
\long\def\@makecaption#1#2{%
  {\centering\normalfont\footnotesize
   #1.\enspace #2\par}%
  \vskip 3pt
}
\makeatother
\caption{\normalfont\textbf{Short-horizon real-world skill evolution.} }
\label{tab:real_short}
\begingroup
\scriptsize
\setlength{\tabcolsep}{4pt}
\renewcommand{\arraystretch}{1.0}
\begin{tabular*}{\textwidth}{@{\extracolsep{\fill}}l*{8}{c}@{}}
\toprule
Task & \multicolumn{2}{c}{Attempts $\downarrow$} & \multicolumn{2}{c}{Output tokens (M) $\downarrow$} & \multicolumn{2}{c}{Total tokens (M) $\downarrow$} & \multicolumn{2}{c}{ Success rate $\uparrow$} \\
\cmidrule(lr){2-3}\cmidrule(lr){4-5}\cmidrule(lr){6-7}\cmidrule(l){8-9}
& w/o Sim & w/ Sim & w/o Sim & w/ Sim & w/o Sim & w/ Sim & w/o Sim & w/ Sim \\
\midrule
Move plate from rack to mat & $4$ & $\mathbf{1}$ & $0.278$ & $\mathbf{0.060}$ & $103.45$ & $\mathbf{17.58}$ & $3/5$ & $\mathbf{4/5}$ \\
Insert straw into cup & $2$ & $\mathbf{1}$ & $0.088$ & $\mathbf{0.069}$ & $\mathbf{22.21}$ & $34.65$ & $2/5$ & $\mathbf{5/5}$ \\
Transfer fried egg from pan onto plate & $2$ & $\mathbf{1}$ & $0.191$  & $\mathbf{0.074}$ & $85.34$  & $\mathbf{24.64}$ & $1/5$ & $\mathbf{4/5}$ \\
\bottomrule
\end{tabular*}
\par\smallskip
\begin{minipage}{\textwidth}
\scriptsize
We compare evolution from empty and frozen simulation libraries by attempts and tokens to first complete-task success and frozen-test successes ($k/5$). w/o Sim: evolution from an empty library; w/ Sim: evolution initialized from the frozen simulation library. M denotes $10^6$ tokens. Bold marks the better result within each pair.
\end{minipage}
\endgroup
\vspace{-3mm}
\end{table*}

\begin{figure*}[!t] \centering \includegraphics[width=\textwidth]{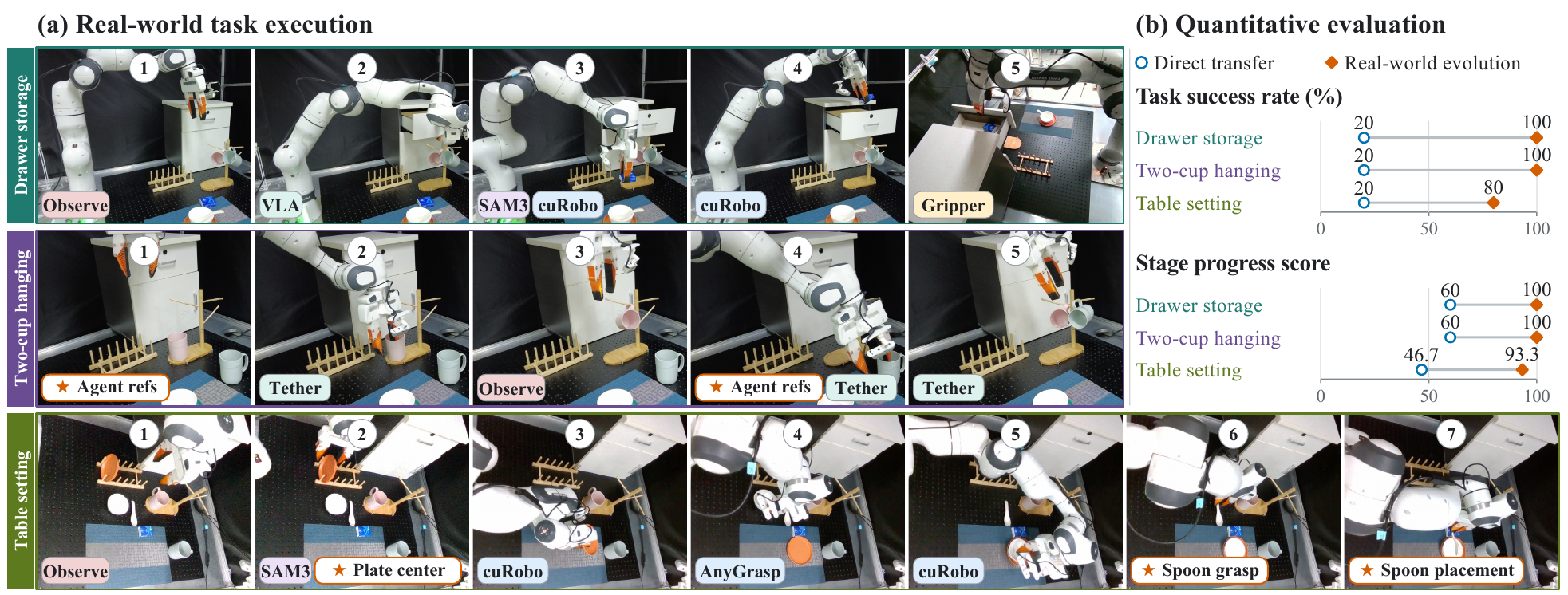}
\vspace{-6mm}
\caption{\textbf{Long-horizon real-world evaluation.}
\textbf{(a)} Real-world-evolution keyframes numbered chronologically. Labels identify tool use, with category colors following Fig.~\ref{fig:pipeline}. \emph{Observe} denotes acquiring current scene images through the observation interface. \textcolor{orange}{$\star$} marks skill-guided, agent-derived references or parameters, including plate centers and spoon grasp or release references. Two-cup hanging uses Tether, absent from the simulation skill library.
\textbf{(b)} \emph{Direct transfer} and \emph{Real-world evolution} start from the same frozen simulation library ($5$ test trials/task). Trial-averaged stage progress ($0$--$100$) equally weights stages: $1/3$ per drawer-storage stage, $1/2$ for each cup hung successfully, and $1/6$ per table-setting grasp or placement.}
  \label{fig:real_long}
\vspace{-12pt}
\end{figure*}

To characterize the tool dependence of the evolved skill library, we independently remove perception, planning, or the VLA tool at test time.
The tool-removal variants retain the full system's frozen skill library, reasoning configuration, control tools, and episode budget. Each condition is evaluated on the same $10$ simulation tasks with $5$ trials each. The agent can replan using remaining tools, while skill updates remain disabled. This protocol evaluates how established execution procedures respond to changes in available capabilities.

\noindent\textbf{Tool removal degrades execution despite replanning.}
Removing any of the three tool groups lowers observed SR on all $10$ tasks (Fig.~\ref{fig:tool_ablation_tasks}(a)). Mean SR decreases from $92.0\%$ for full EvoTER to $50.0\%$ without perception tools and $40.0\%$ without planning tools, corresponding to reductions of $42.0$ and $52.0$~pp, respectively. Among the three tool-group ablations, VLA removal produces the largest average decrease, reducing mean SR to $28.0\%$. Although the agent can replan, the remaining tools do not fully compensate for the removed capabilities under the existing skill library. These results show that reusable procedural knowledge depends on the available execution capabilities.

\noindent\textbf{The dominant tool dependence varies across tasks.}
The aggregate ranking masks substantial differences between tasks (Fig.~\ref{fig:tool_ablation_tasks}(a)). On \mbox{RecycleBottlesByType}, removing planning tools reduces SR to $0\%$, versus $80\%$ SR without VLA. On \mbox{StirVegetables}, the ordering reverses: SR falls to $0\%$ without VLA but remains $20\%$ without planning tools. Thus, the largest average loss from VLA removal coexists with tasks that depend more strongly on planning. The evolved agent shows task-specific reliance on available capabilities, requiring per-task analysis of aggregate ablations.

\subsection{Skill Transfer and Evolution in the Real World}
\label{sec:real}
\noindent\textbf{Evaluation Protocol.}
Our evaluation examines whether simulation-acquired skills facilitate real-world development and how real-world experience refines transferred procedures. Short-horizon experiments compare skill evolution with and without simulation initialization (Table~\ref{tab:real_short}). Long-horizon experiments examine direct transfer and subsequent evolution from the same frozen simulation library (Fig.~\ref{fig:real_long}).

\noindent\textbf{Simulation Initialization for Short-Horizon Evolution.}
We compare evolution from an empty library (w/o Sim) with evolution initialized from the frozen simulation library (w/ Sim). The evaluation comprises $3$ tasks with related procedures in the simulation library: moving a plate from a rack to a mat, inserting a straw into a cup, and transferring a fried egg from a pan onto a plate. Development is capped at $150$~M total tokens per task. We measure attempts and cumulative output and total token counts until the first complete success, then freeze the resulting library and evaluate it in $5$ independent trials with different initial layouts.

Simulation initialization reduces development attempts on all three tasks. Each task reaches its first complete success in one attempt,
compared with $4$, $2$, and $2$ attempts without simulation initialization for plate transfer, straw insertion, and fried egg transfer, respectively.
The benefit extends to subsequent frozen execution: success counts increase from $3/5$ to $4/5$, from $2/5$ to $5/5$, and from $1/5$ to $4/5$. These results support reusing simulation-acquired procedures as prior knowledge for both real-world skill development and execution under varied initial layouts.

The acquisition benefit depends on the cost metric. Simulation initialization reduces output-token consumption on all three tasks, but total-token costs vary. Plate transfer decreases from $103.45$~M to $17.58$~M total tokens, and fried egg transfer from $85.34$~M to $24.64$~M. For straw insertion, total-token consumption increases from $22.21$~M to $34.65$~M despite requiring fewer attempts. Thus, fewer development attempts do not always translate into lower total-token cost.

\noindent\textbf{Real-World Feedback for Long-Horizon Skill Evolution.}
To assess the contribution of real-world feedback to long-horizon execution, we compare \emph{Direct transfer} with \emph{Real-world evolution} on table setting, drawer storage, and two-cup hanging. Table setting arranges a plate, bowl, and spoon, with later placements accounting for the actual plate and bowl positions. Drawer storage requires opening the drawer before placing a tissue packet inside. Two-cup hanging requires hanging both cups on a rack and examines how the skill library develops procedures beyond those acquired in simulation. For this task, Tether provides the hanging capability and is available to both conditions before comparison, with its implementation fixed. Both conditions start from the same frozen simulation library, independently of short-horizon development. \emph{Direct transfer} keeps the library unchanged, whereas \emph{Real-world evolution} revises its tool-use procedures through real-world development before freezing. Both permit observation, replanning, and recovery per test episode under the same execution budget, without library updates between test episodes. Success requires all task-specific goal relations to hold at episode end. Long-horizon evolution from an empty library is not evaluated within the real-robot development budget.

\noindent\textbf{Real-world experience refines execution procedures.}
Following real-world evolution, the agent completes table setting in $4/5$ trials, drawer storage in $5/5$ trials, and two-cup hanging in $5/5$ trials
(Fig.~\ref{fig:real_long}(b)), compared with $1/5$ on each task under direct transfer. For table setting under direct transfer, the five trials complete $4,0,6,4,0$ of six stages, respectively, yielding mean stage progress of $100\times14/30\approx46.7$. The revisions illustrate how real-world experience updates both state-dependent manipulation rules and procedures for invoking newly available tools. For example, a revised table-setting rule computes the spoon release height from the current bowl geometry and held-spoon pose.
Tether warps a demonstration trajectory using visually matched grasp and release references and executes it open-loop~\cite{liang2026tether}. Native invocations repeatedly failed because the matched references deviated from the intended actions. After diagnosing these failures and receiving a high-level hint to select the references itself, the agent evolved a skill that selects them from the observed cup, handle, and rack and revises those rejected by cross-view triangulation, while Tether remains unchanged.

\section{CONCLUSION}
\label{sec:conclusion}
We presented EvoTER, an experience-driven framework evolving a persistent tool-use skill library with model parameters and tool implementations fixed within each phase.
Its Explorer--Diagnoser--Validator loop retains revisions only when matched complete-task evaluations in fresh sessions show aggregate success gains over the current library.
Validated simulation skills initialize real-world evolution guided by physical feedback under changing conditions and tools.
In simulation, EvoTER achieves higher overall success than representative policy-based and agent-based baselines, with matched evaluations showing gains over its pre-evolution baseline, particularly on composite tasks.
On real robots, simulation initialization reduces short-horizon acquisition attempts, while continued evolution substantially improves long-horizon success over direct transfer and refines procedures for newly available tools.
These findings support validated procedural knowledge for continual adaptation without model retraining.
Future work will streamline validation, reduce reliance on privileged simulation feedback, and broaden evaluation across tasks and embodiments.

\bibliographystyle{IEEEtran}
\bibliography{reference}

\end{document}